\documentclass[letterpaper, 10 pt, conference]{ieeeconf}  

\IEEEoverridecommandlockouts                              

\usepackage{graphics} 
\usepackage{epsfig} 
\usepackage{amsmath} 
\usepackage{amssymb}  
\usepackage{cite}
\usepackage{multirow}
\usepackage{booktabs}
\usepackage{xcolor}
\usepackage{hyperref}
\usepackage[normalem]{ulem}

\DeclareMathAlphabet{\mathpzc}{T1}{pzc}{m}{it}

\definecolor{antiquefuchsia}{rgb}{0.57, 0.36, 0.51}

\renewcommand\bf[1]{\textbf{#1}}
\newcommand\ul[1]{\underline{#1}}
\newcommand\uul[1]{\underline{\underline{#1}}}

\usepackage{mathtools}

\DeclarePairedDelimiter\floor{\lfloor}{\rfloor}

\usepackage{amsfonts}
\usepackage{stackengine}
\stackMath

\title{\LARGE \bf {Joint Object Detection and Multi-Object Tracking with Graph Neural Networks}
\vspace{-0.2cm}}

\author{Yongxin Wang, Kris Kitani, Xinshuo Weng 
\vspace{-1cm}
\thanks{All authors are with the Robotics Institute,
        Carnegie Mellon University. {\tt\small \{yongxinw, kkitani, xinshuow\}@cs.cmu.edu}}
        }%

\begin{document}

\maketitle
\thispagestyle{empty}
\pagestyle{empty}

\begin{abstract}

Object detection and data association are critical components in multi-object tracking (MOT) systems. Despite the fact that the two components are dependent on each other, prior works often design detection and data association modules separately which are trained with separate objectives. As a result, one cannot back-propagate the gradients and optimize the entire MOT system, which leads to sub-optimal performance. To address this issue, recent works simultaneously optimize detection and data association modules under a joint MOT framework, which has shown improved performance in both modules. In this work, we propose a new instance of joint MOT approach based on Graph Neural Networks (GNNs). The key idea is that GNNs can model relations between variable-sized objects in both the spatial and temporal domains, which is essential for learning discriminative features for detection and data association. Through extensive experiments on the MOT15/16/17/20 datasets, we demonstrate the effectiveness of our GNN-based joint MOT approach and show state-of-the-art performance for both detection and MOT tasks. Our code is available at: \textcolor{blue}{\url{https://github.com/yongxinw/GSDT}}.

\end{abstract}

\section{Introduction \label{sec:intro}}

Object detection \cite{Girshick2015, Ren2015, Zagoruyko2016, Redmon2016, Lin2017, Singh2018, Weng2019, He2019} and data association \cite{Xiang2015, Bewley2016, Wojke2017, Schulter2017, Maksai2017, Weng2020_AB3DMOT, Karunasekera2019, Wang20192} are two components of multi-object tracking (MOT), which is essential to perception in robotic systems such as autonomous driving \cite{Luo2018, Wang2018, Yurtsever2019, Badue2019, Weng2020_SPF2}. Prior work \cite{Bewley2016, Weng2020_AB3DMOT} often approaches MOT in an online fashion using a tracking-by-detection pipeline, where a detector outputs detections followed by a data association module matching the detections with past tracklets to form new tracklets up to the current frame. Oftentimes, the detector and data association modules are trained separately in prior work. However, with this separate optimization procedure, we cannot back-propagate errors through the entire MOT system. In other words, each module is optimized only towards its own local optimum, but not towards the objective of MOT. As a result, this separate optimization procedure used in prior work often yields sub-optimal performance. 

To improve performance, we investigate 1) joint optimization of object detection and data association, which we refer to as the joint MOT framework; 2) within the joint MOT framework, how to learn more discriminative features. First, to address the joint MOT problem, prior work \cite{B2016, Feichtenhofer2017, Bergmann2019, zhou2020tracking, Voigtlaender2019, Wang2019, zhang2020fairmot, peng2020chainedtracker, pang2020tubetk, Sun2020} has explored different directions. \cite{B2016, Bergmann2019, Feichtenhofer2017, zhou2020tracking} proposed to unify object detector with a model-free single-object tracker, where the tracker directly regresses the location of each object detected in the previous frame to the current frame. As each object is tracked independently, the data association problem is naturally resolved. \cite{Voigtlaender2019, Wang2019, zhang2020fairmot, peng2020chainedtracker} proposed to extend an object detector by adding a re-identification (Re-ID) \cite{cuhk} or ID verification branch which extracts features of objects for matching across frames. Also, \cite{pang2020tubetk, Sun2020} proposed an anchor tube. Different from anchor box used in anchor-based detectors, anchor tube represents a sequence of bounding boxes in a list of frames (one box per frame). Given video clips as inputs, true positive tubes can be found and used as tracklet outputs, which solves the joint MOT problem in a single shot. 

\begin{figure}[t]
    \vspace{0.17cm}
    \centering
    \includegraphics[trim=0.1cm 0.3cm 0.3cm 0.1cm, clip=true, width=\linewidth]{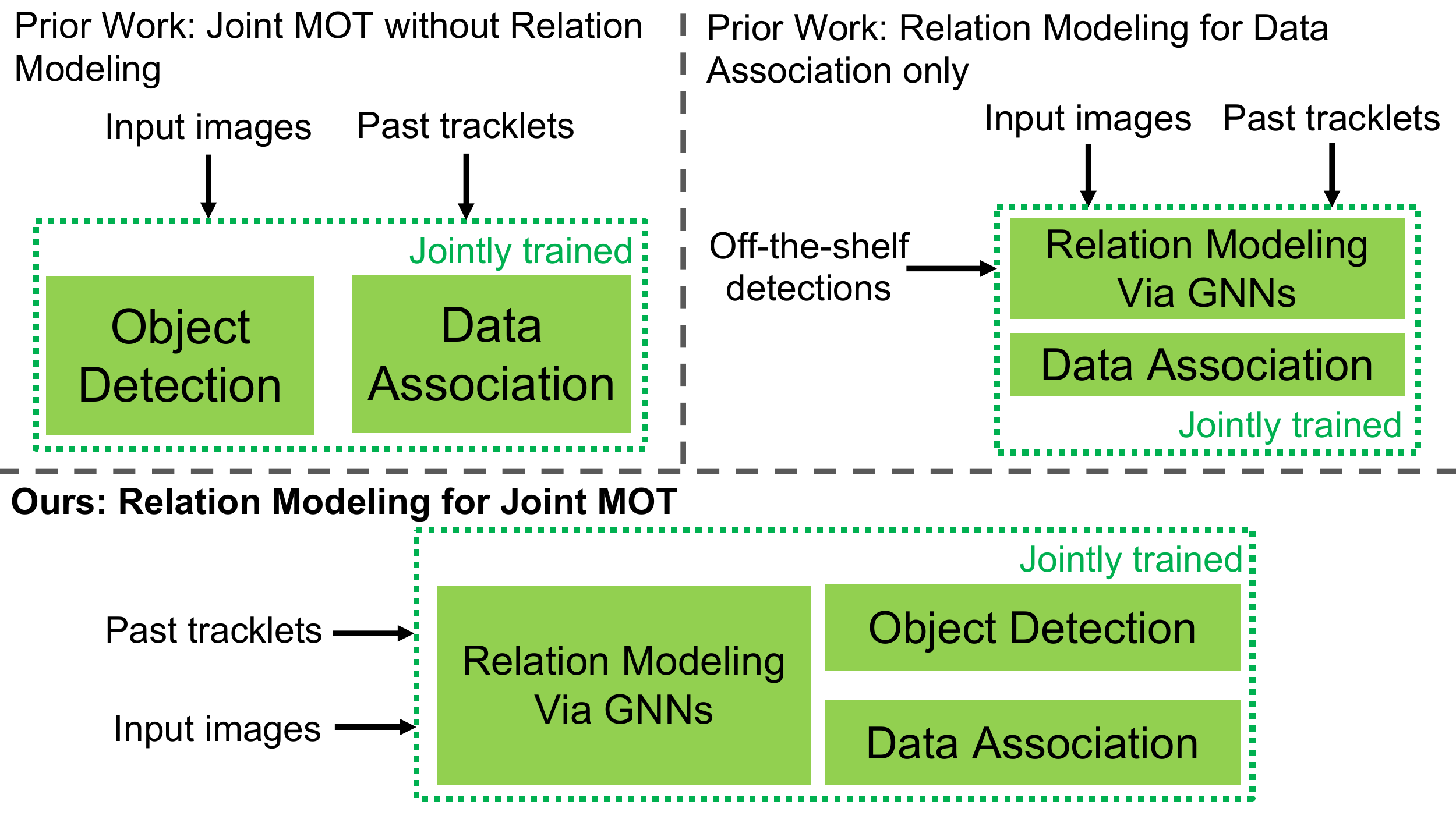}
    \vspace{-0.75cm}
    \caption{(\bf{Top left}): Although jointly training detection and data association, prior work does not take into account object-object relations. (\bf{Top right}): Prior work leverages object-object relations but only adopts it for data association. By using off-the-shelf detections that cannot be optimized jointly, such disjoint MOT paradigm can lead to sub-optimal MOT performance. (\bf{Bottom}): Our method leverages spatial-temporal object relations for both detection and data association under a joint MOT framework.}
    \label{fig:teaser}
    \vspace{-0.5cm}
\end{figure}

Although prior joint MOT methods achieved impressive performance, we observed that the feature extracted for individual object/tracklet/tube is independent of one another and object-object relations are ignored. We argue that such object relations are useful to both object detection and data association. For example, for object detection in MOT, if related objects (\emph{e.g.}, two pedestrians that walk together) co-occur in the last frame, it is likely that they will also co-occur in the current frame at a nearby location. For data association, if the similarity score of two objects across frames increases with a high confidence (\emph{i.e.}, it is likely that these two objects have the same identity), then the similarity score between any of these two objects and other objects should be suppressed to avoid confusion in data association. As a result, to achieve better performance for the joint MOT framework, we should design our method to leverage object-object relations. While recent works \cite{Weng2020_gnn3dmot, Braso2020, Li2020_gn} have exploited object-object relations in data association, they are limited to disjoint MOT where the detector is optimized separately. Therefore, the object relations are not used for the detector which is sub-optimal.

To deal with the above issue, we propose a new instance of joint MOT approach that can model object-object relations for both object detection and data association. Specifically, to obtain more discriminative features, we use Graph Neural Networks (GNNs) to exploit object-object relations. Then, the obtained features that consider object relations are used for both tasks of detection and data association. With GNNs, the feature extracted for each object is not isolated anymore, but instead can be adapted via features of its related objects in both spatial and temporal domains. To the best of our knowledge, our work is the first to model object relations via GNNs in the joint MOT framework. Our work overcomes the limit of prior work that 1) can perform joint MOT while ignoring object relations, or 2) can model object relations in association but cannot address joint MOT and use object relations to improve detection. Through experiments on MOTChallenges, we show S.O.T.A. performance on MOT15/20 and competitive performance on MOT16/17 datasets. Our contributions are summarized as follows. 
\begin{enumerate}
    \item A joint MOT approach that models object relations via GNNs to improve both detection and data association;
    \item Strong empirical performance on MOT 15/16/17/20 datasets for both detection and MOT tasks.
\end{enumerate}

\section{Related Works}

\noindent\textbf{Object Detection.} There have been tremendous advancements in image-based detection since large datasets such as COCO \cite{Lin2014} have been introduced. In past few years, anchor-based object detectors \cite{Ren2015, Liu2016, He2017, Redmon2016} have been dominant in the field of object detection. Recently, a new type of detector that models objects as points has been proposed \cite{Zhou2019, Law2018} and achieved impressive performance. However, image-based methods suffer from unstable detections across frames as they use a single image as input.  To deal with the issue, video  detection \cite{Zhu2018, Ramzy2019, Xiao2018} has been investigated which uses multiple frames as inputs. Although producing more temporally-consistent detections than image-based methods, existing video-based methods cannot model object relations. Similar to video-based detection methods, our joint MOT method also takes multiple frames as inputs to obtain detections. However,  our difference is that, in additional to leveraging multiple frames of input, we also model object relations via GNNs, which can further improve detection performance when the objects across consecutive frames are highly related to one another. (\emph{e.g.}, object co-occurring.

\vspace{2mm}\noindent\textbf{Multi-Object Tracking.} Recent MOT work primarily focuses on data association component in the tracking-by-detection pipeline, which can be split into online and batch methods. Online methods \cite{Bewley2016, Weng2020_AB3DMOT, Zhang2019, Zhu2018_MOT} only require information up to the current frame for data association and can be useful to online applications. On the other hand, batch methods \cite{Milan2014, Pirsiavash2011, B2016_mot, Brendel2011, Dehghan2015, RoshanZamir2012} need to have access to global information (\emph{i.e.}, information up to the current frame plus information from future frames), which can theoretically achieve higher accuracy than online methods but are not applicable to online scenarios. We restrict the scope of this paper to online methods.  Different from prior online methods that use off-the-shelf detections and only focus on the data association component, our method jointly optimizes detection and data association with an additional GNN module for object-object relation modeling which improves performance in both tasks. 

\vspace{2mm}\noindent\textbf{Joint Detection and Data Association.} To enable gradient back-propagation in the joint MOT framework and improve overall performance, \cite{B2016, Feichtenhofer2017, Bergmann2019, zhou2020tracking, Voigtlaender2019, Wang2019, zhang2020fairmot, peng2020chainedtracker, pang2020tubetk, Sun2020} attempted to jointly optimize detection and data association for MOT. As introduced before, prior joint MOT methods can be mostly split into three categories: 1) unify a single-object tracker with an object detector; 2) add a Re-ID branch to object detection network; 3) use anchor tube representation for one-shot joint MOT. For example, \cite{Bergmann2019} builds on top of Faster-RCNN \cite{Ren2015} and makes its bounding box regression head multi-purpose, \emph{i.e.}, not only to refine the detected bounding boxes but also to serve as a single object tracker that regresses the object location from the previous frame to the current frame. Our method shares a similar spirit with prior joint MOT methods but goes beyond them by modeling spatial-temporal object relations. We show that our method can improve performance for both object detection and data association tasks in MOT.

\vspace{2mm}\noindent\textbf{Graph Neural Networks for Relation Modeling.} GNNs were first introduced by \cite{Gori2005} to process data with a graph structure using neural networks. The key idea is to construct a graph with nodes and edges relating each other and update node/edge features based on relations, \emph{i.e.,} a process called node feature aggregation. In recent years, different GNNs (\emph{e.g.}, GraphConv~\cite{Morris2019}, GCN \cite{Kipf2017}, GAT \cite{Velickovic2018}, \emph{etc}) have been proposed each with a unique feature aggregation rule which are shown to be effective on various tasks. Specifically for MOT, recent work \cite{Braso2020, Weng2020_gnn3dmot, Li2020_gn, Weng2020_forecast} formulates data association as an edge classification problem using GNNs, where each node represents an object and each edge relating two nodes denotes the similarity between detection and tracklets. These approaches use GNNs to update node features via object relation modeling and have shown improved MOT performance. As opposed to these methods that focus only on using GNNs to improve data association, our work uses GNNs for joint detection and association, where the detection branch also benefits from the GNN relation modeling. 

\begin{figure*}[t]
    \vspace{1.5mm}
    \centering
    \includegraphics[width=\linewidth]{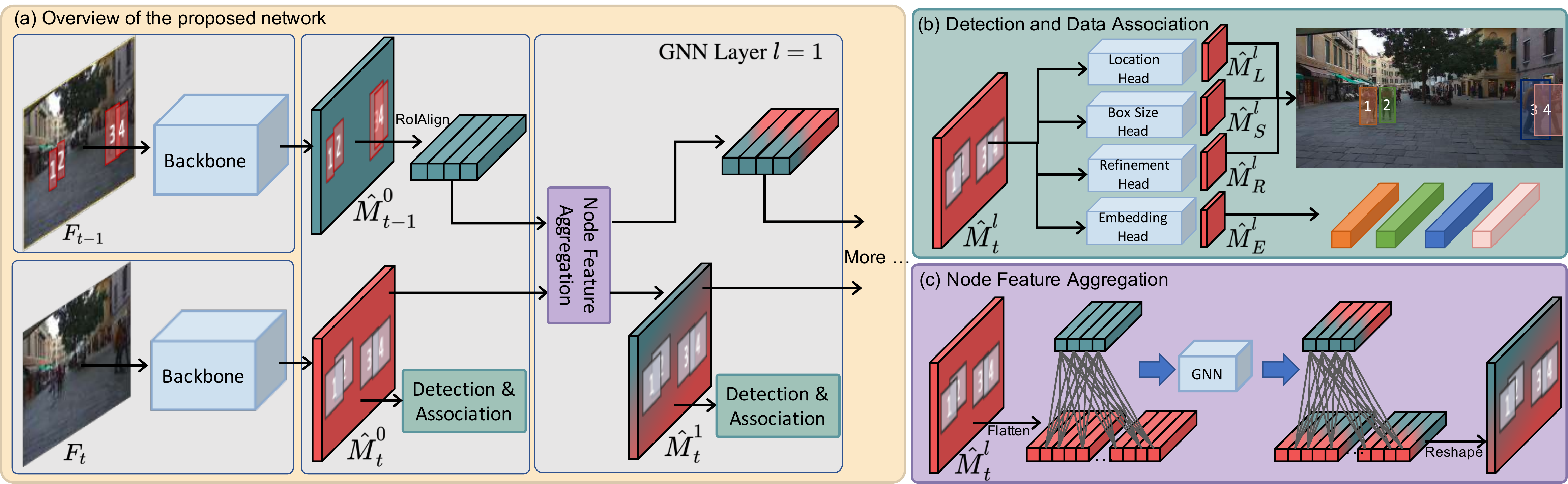}
    \vspace{-0.75cm}
    \caption{\textbf{(a) Overview of the Proposed Network}. We first extract features $\hat{M}_{t-1}^0$, $\hat{M}_{t}^0$ from images $F_t$ and $F_{t-1}$ using a shared backbone. To obtain feature of each tracklet in $T_{t-1}$, we use RoIAlign to crop feature from the image feature $\hat{M}_{t-1}^0$ given the tracklets' boxes (red boxes in $\hat{M}_{t-1}^0$). To obtain features of potential detections, we use feature of every pixel in $\hat{M}_{t}^0$. To construct a graph with manageable number of edges, we only define edges between features of potential detections to tracklets if their spatial distances are within a window (grey boxes in $\hat{M}_{t}^0$). With the constructed graph, we use 3-layer GNNs to update features of tracklets and potential detections via node feature aggregation. A detection and data association head is applied to every layer of GNNs to obtain final detections and matching. \textbf{(b) Detection and Data Association}: The location, box size, and refinement heads generate $\hat{M}^l_L$, $\hat{M}^l_S$, and $\hat{M}^l_R$ which are used to obtain detections. The embedding head generates $\hat{M}^l_E$ to compute identity embedding for data association. \textbf{(c) Node Feature Aggregation}. The mixed color illustrates that features from tracklets and potential detections affect each other via relation modeling.}
    \label{fig:architecture}
    \vspace{-0.35cm}
\end{figure*}

\section{Approach}

We follow online MOT methodology. Given images $F_{t-1}$, $F_t$ at frame $t$-$1$, $t$ and tracklets at frame $t$-$1$, denoted as $T_{t-1}$=$\{T_{t-1}^1, T_{t-1}^2, ..., T_{t-1}^{P_{t-1}}\}$ as inputs, we aim to detect objects $D_t = \{D_t^1, D_t^2, ..., D_t^{K_t}\}$ in $F_t$, and associate them to tracklets $T_{t-1}$, where $K_t$ and $P_{t-1}$ are number of detections and tracklets which can vary across frames. By associating detections to past tracklets, we can determine whether to extend or terminate an existing tracklet or to initiate a new tracklet at frame $t$. We perform this process iteratively at every frame to obtain the tracklets of the entire video.

To simultaneously detect and associate objects for MOT, we integrate a detector and a Re-ID module in our model. However, doing so alone does not leverage spatial-temporal object relations. Therefore, we use GNNs to extract relations between objects and learn better features to improve both detection and data association. For illustration, we show the proposed network architecture in Fig. \ref{fig:architecture}, which contains four modules: feature extraction (Sec. \ref{sec:det}), node feature aggregation (Sec. \ref{sec:gnn}), detection (Sec. \ref{sec:det}) and association (Sec. \ref{sec:asso}). As we use \underline{\textbf{G}}NNs for \underline{\textbf{S}}imultaneous \underline{\textbf{D}}etection and \underline{\textbf{T}}racking, we abbreviate our method as GSDT.

\subsection{Feature Extraction and Object Detection \label{sec:det}}
Given two input images $F_{t-1}$ and $F_t$, we first use a shared DLA-34 backbone \cite{yu2019deep} to extract feature maps at two frames $\hat{M}^0_{t-1}, \hat{M}^0_t \in \mathbb{R}^{\frac{W}{r} \times \frac{H}{r} \times C}$ as shown in Fig. \ref{fig:architecture} (a) left, where $r$ is the downsample ratio, $W$/$H$ are width and heights of the images, and $C$ is the number of channels. Both image features $\hat{M}^0_{t-1}, \hat{M}^0_t$ will be used in the following relation modeling, object detection and data association modules. Also, for the image feature at frame $t$, we will update it at the $l$th layer of GNNs to obtain $\hat{M}^l_t$ (Sec. \ref{sec:gnn}).

For object detection, we follow CenterNet \cite{centerNet} and detect each object by finding its center coordinate $(x, y)$ and width/height $(w, h)$. As we only perform detection in the frame $t$, we feed $\hat{M}^l_t$ to three heads (location, box size and refinement heads) as shown in Fig. \ref{fig:architecture} (b), which provides three maps, \emph{i.e.}, a location map $\hat{M}^l_L \in \mathbb{R}^{\frac{W}{r} \times \frac{H}{r}}$, a size map $\hat{M}^l_S \in \mathbb{R}^{\frac{W}{r} \times \frac{H}{r} \times 2}$, and a refinement map $\hat{M}^l_R \in \mathbb{R}^{\frac{W}{r} \times \frac{H}{r} \times 2}$. As the names suggest, $\hat{M}^l_L$ provides rough estimates of object center coordinates, $\hat{M}^l_S$ gives estimated widths and heights, and $\hat{M}^l_R$ refines rough coordinates to be precise. Note that these three detection heads can operate on any GNN layer $l$.

\vspace{1mm}\noindent\textbf{Training}. To train three detection heads and back-propagate the gradients through the entire network, we need to construct ground truth (GT) for each map and apply loss functions. For GT location map $M_L^l \in \mathbb{R}^{\frac{W}{r} \times \frac{H}{r}}$, we use a Gaussian heatmap where the peaks occupy the locations of GT objects.
Specifically, the value at the location $(i, j)$ is: 
\vspace{-0.1cm}
\begin{equation}
\resizebox{0.9\linewidth}{!}{
    $M_L^l(i, j) = \sideset{}{^N_{k=1}} \sum \exp{(-\frac{(i - \floor{\frac{x_k}{r}})^2 + (j - \floor{\frac{y_k}{r}})^2}{2\sigma_k^2})} 
    $},
    \vspace{-0.1cm}
\end{equation}
where $N$ is number of GT objects and $(x_k, y_k)$ is the center coordinate of GT object $k$, and $\floor{\cdot}$ is a floor function. The standard deviation $\sigma_k$ scales with the GT box size $(w_k, h_k)$ \cite{Law2018}. For GT box size and refinement maps $M_S^l, M_R^l$, if there exists a GT object center at location $(i, j)$, then:
\vspace{-0.1cm}
\begin{equation}
    M_S^l(i, j, :) = (w, h),
    \vspace{-0.1cm}
\end{equation}
\begin{equation}
    M_R^l(i, j, :) = (\frac{x}{r} - \floor{\frac{x}{r}}, \frac{y}{r} - \floor{\frac{y}{r}}).
    \vspace{-0.1cm}
\end{equation}

After we construct the GT for three maps, we use focal loss for the location map $L_{loc}^l$, and L1 loss for the size map $L_{size}^l$ and refinement map $L_{ref}^l$. Note that for size/refinement maps, their losses are only applied to locations with GT objects. The overall loss for detection is then:
\vspace{-0.1cm}
\begin{equation}
    L_{det}^l = \lambda_1 L_{loc}^l + \lambda_2 L_{size}^l + \lambda_3 L_{ref}^l,
    \vspace{-0.1cm}
\end{equation}
where we empirically found $\lambda_1 = \lambda_3 = 1, \lambda_2=0.1$ could achieve a well balance between each individual loss.

\subsection{Data Association \label{sec:asso}}

To match detections with tracklets, we add an additional embedding head as shown in Fig. \ref{fig:architecture} (b) to learn an identity embedding for each potential detection, \emph{i.e.}, every pixel in $\hat{M}_t^l$. These identity embeddings are used to compute similarity between every pair of tracklet and potential detection during testing, where the identity embeddings of tracklets are cached when they are detected in the previous frame. Specifically, given $\hat{M}_t^l$ as inputs, our embedding head outputs an embedding map $\hat{M}_E^l  \in \mathbb{R}^{\frac{W}{r} \times \frac{H}{r} \times D}$, where $D$ denotes the embedding dimension and each pixel in $\hat{M}_E^l$ has an embedding for a potential detection centered at this pixel.

\vspace{1mm}\noindent\textbf{Training.} To optimize our embedding head, we further use embedding at each pixel of the embedding map $\hat{M}_E^l$ as input, and predict its identity vector $\hat{\mathbf{p}}$, the $i$th entry of which represents the probability of this pixel has an object identity of $i$. Also, we will need the GT identity vector $\mathbf{p}$ for training, which is a one-hot vector of length $M$ ($M$ is the number of object identities in the target dataset). Note that in the GT identity vector $\mathbf{p}$, only the GT identity index is filled with $1$ while all other indexes are $0$s. Similar as before, we only supervise our embedding map $\hat{M}_E^l$ at pixels where there exists a GT object using a cross-entropy loss as follows:
\vspace{-0.1cm}
\begin{equation}
    L_{emb}^l = \frac{1}{N} \sideset{}{_{k=1}^N} \sum \sideset{}{_{m=1}^{M}} \sum \mathbf{p}^k(m)log(\hat{\mathbf{p}}^k(m)) .
\end{equation}

\subsection{Graph Neural Networks for Relation Modeling \label{sec:gnn}}
As detection and data association modules alone overlook spatial-temporal relations between tracklets and detections, we use GNNs to further leverage such relation information and obtain better features for joint MOT.

\vspace{1mm}\noindent\textbf{Graph Construction.} To perform graph feature learning, we need to construct a graph $G(V, E)$ with nodes $V$ being features of detections and tracklets. For features of tracklets, we can easily obtain them as tracklets' boxes are given as inputs. Specifically, we crop tracklet features from the image feature $\hat{M}_{t-1}^0$ using the RoIAlign \cite{He2017} operator. However, it is relatively non-trivial to obtain features of detections. This is because objects have not been detected in $\hat{M}_t^l$ so we do not have their locations to crop features from $\hat{M}_t^l$. To overcome this issue, we use feature at every pixel of $\hat{M}_{t}^l$ to represent a potential detection, resulting in $\frac{W}{r} \times \frac{H}{r}$ detection nodes. This design choice is made because we believe that $\hat{M}_{t}^l$ contains enough information about detections, as all three detection heads are branched from $\hat{M}_{t}^l$. 

In addition to node construction, we also need the set of edges $E$. A simple solution is to define an edge between every pair of nodes, resulting in a fully-connected graph. However, such a graph with massive number of edges can be computationally expensive and sometimes impractical when the number of tracklets and potential detections (\emph{i.e.}, number of pixels in $\hat{M}_{t}^l$) are too large. Instead, we take advantage of domain knowledge in MOT: (a) data association only happens across frames (not in the same frame); (b) displacement of the same object is often small across frames. According to (a), we only define edges between tracklet nodes and detection nodes. Based on (b), for every tracklet node, we only connect it to detection nodes that are in nearby locations, \emph{i.e.}, within a $s \times s$ spatial window centered at the tracklet location in $F_{t-1}$. We use $s=15$ in our model which balances computation and performance. We illustrate the idea of edge construction in Fig. \ref{fig:architecture}, where grey boxes on $\hat{M}^l_{t}$ enclose the pixels (detection nodes) that have edges with tracklets.

\vspace{1mm}\noindent\textbf{Node Feature Aggregation.} The key idea of our method is using GNNs to model object-object relations and improve feature learning for both detection and data association. To that end, we iteratively update node features by aggregating features from its neighborhood nodes connected with edges. In this way, information can propagate through the graph and nodes are allowed to interact. So the question is how exactly we should perform the node feature aggregation? 

Recently, there are many GNNs proposed, \emph{e.g.}, GraphConv \cite{Morris2019}, AGNNConv \cite{thekumparampil2018attentionbased}, EdgeConv \cite{Wang2019_graph}, each with a different node aggregation rule. To explore how different GNNs affect our joint MOT method, we try five popular GNNs in our model in the ablation study and compare their performance. Our final model uses GraphConv as it provides the best empirical performance. Specifically, GraphConv updates node features using the following rule:
\vspace{-0.1cm}
\begin{equation}
    h^i_l = \rho_1 (h^i_{l-1}) + \sideset{}{_{j \in \mathpzc{N}(i)}}\sum \rho_2 (h^j_{l-1}), \label{eq:graphconv}
    \vspace{-0.1cm}
\end{equation}
where $h^i_l$ represents the feature of node $i$ at GNN layer $l$, $\mathpzc{N}(i)$ defines neighborhood of node $i$, and $\rho_1, \rho_2$ are linear layers. Note that, for detection nodes that have no edge to any tracklets, the second term in Eq. \ref{eq:graphconv} disappears, and these potential detection nodes are only used for discovering new objects in frame $t$ (not having corresponding tracklet in frame $t$-$1$). Also, one might argue that, as our edges are only defined across frames, object-object relations are only modeled in the temporal domain for data association but not in the spatial domain for object detection, \emph{i.e.}, nodes within the same frame cannot interact. This is true if we only have a single layer of GNNs. In the case we use more than one layer, the information can be propagated back and forth, enabling spatial-temporal object relation modeling. For example, in the first layer, the feature of a detection node $i_1$ is aggregated to a tracklet node $j$. Then in the second layer, the feature of the tracklet node $j$ is propagated to a different detection node $i_2$, \emph{i.e.}, part of the node feature $i_1$ is propagated to the detection node $i_2$ which enables spatial relation modeling in the same frame. In ablation study, we will explore how the number of GNN layers will affect performance of our model. In our best model, we use three layers of GNNs.

\subsection{Joint Detection and Association with GNNs}
Although the detection and association heads introduced in Sec. \ref{sec:det} and \ref{sec:asso} can solve joint detection and data association problem by training two modules together, they do not leverage object relations if they only use features from $\hat{M}_{t-1}^0, \hat{M}_{t}^0$. As shown in Fig. \ref{fig:architecture}, to leverage GNNs, we apply the detection and association head also to features obtained in GNN layer 1, 2, \emph{etc}, which have better features after node feature aggregation by encoding relations. To summarize, the overall loss of our network is a summation of detection and data association losses over all layers of GNNs:
\vspace{-0.1cm}
\begin{equation}
    L_{total} = \sideset{}{_l}\sum \eta_1 L_{det}^l + \eta_2 L_{emb}^l,
    \vspace{-0.1cm}
\end{equation}
where $l$ is the index of GNN layers and $\eta_1, \eta_2$ denote weights of each loss. We adopt an automatic loss weighting scheme as in \cite{8578879} to balance $\eta_1$ and $\eta_2$. We refer our readers to \cite{8578879} for details about this automatic loss weighting.

\subsection{Inference and Tracking Management}

At test time, we iteratively detect objects and associate them to existing tracklets. At every frame, we first obtain estimated detections $D_t$ in $F_t$ by post processing $\hat{M}_L^l$, $\hat{M}_S^l$, and $\hat{M}_R^l$ followed by non-maximal suppression. Also, we obtain identity embeddings from $\hat{M}_E^l$ at pixels where objects are detected, which provides us identity embeddings of detections $D_t$. The identity embeddings for $T_{t-1}$ are cached when they are detected in the previous frame so they are also available. We use these embeddings to compute affinity matrix $\hat{\mathbf{S}}$ where each entry represents the similarity between every pair of tracklet and detection. We then feed $\hat{\mathbf{S}}$ to the Hungarian Algorithm \cite{WKuhn1955} to find the best matches between $D_t$ and $T_{t-1}$. For each detection that is not matched to any tracklets, we use it to initialize a new tracklet if its detection confidence score is higher than $0.4$. For unmatched tracklets, they might either exit the scene or be still in the scene but not detected. To avoid terminating tracklets that are still in the scene, we use the Kalman filter \cite{Kalman1960} to extend the unmatched tracklets for $\tau_{term}$ frames, while keep trying to match these tracklets with detections in following frames. If after $\tau_{term}$ frames an unmatched tracklet still cannot be matched to any new detection, we terminate this tracklet. Empirically, we found $\tau_{term}=30$ works well in our model.

\section{Experiments}

\noindent\textbf{Datasets.} We evaluate our method on MOTChallenges \cite{Taixe2015, Milan2016, Dendorfer2019, Dendorfer2020}, including the 2DMOT15/MOT16/MOT17/MOT20 benchmarks. These datasets have two tracks: public and private, where the public track asks methods to use off-the-shelf detections provided by the datasets while methods in the private track can use their own detections. As we jointly detect and track objects (not using off-the-shelf detections for disjoint MOT), it is straightforward to evaluate our method in private track. For a fair comparison with prior work, we evaluate on the test set by submitting our results to the MOT test server.

\vspace{1mm}\noindent\textbf{Evaluation Metrics.} For MOT, we use standard CLEAR MOT metrics \cite{Bernardin2008} and IDF1 metric \cite{ristani2016performance}. For object detection, we report the Average Precision (AP) using official MOT17Det and MOT20Det evaluation protocol. 

\vspace{1mm}\noindent\textbf{Implementation Details.}
We use an Adam optimizer with an initial learning rate of $1e^{-4}$ decaying by a factor of $0.1$ at epoch $20$ and $27$. We train the network for a total of $30$ epochs with a batch size of 16. Following TR-MOT \cite{Wang2019}, we use a mix of six pedestrian detection datasets for pre-training, including CalTech\cite{caltech}, MOT17\cite{Milan2016}, CUHK-SYSU\cite{cuhk}, ETH\cite{eth}, PRW\cite{zheng2017person}, and CityPersons\cite{8099957}. Same as \cite{Wang2019}, we removed a few sequences in the above six datasets which overlap with MOT test set to avoid training with test data, for the sake of fair comparison.

\begin{table}[t]
\vspace{1.5mm}
\setlength\tabcolsep{2pt}
\centering
\caption{MOT evaluation on 2DMOT2015/MOT16/MOT17/MOT20 test sets as of 2021/03/22 (published methods only).}
\vspace{-0.35cm}
\begin{tabular}{@{}llrrrrr@{}}
\toprule
& Method & MOTA(\%)$\uparrow$ & IDF1(\%)$\uparrow$ & MT(\%)$\uparrow$ & ML(\%)$\downarrow$ & IDS$\downarrow$\\
\midrule

\parbox{2mm}{\multirow{10}{*}{\rotatebox[origin=c]{90}{2DMOT2015}}} 
& DMT \cite{B2016} & 44.5 & 49.2 & 34.7 & 22.1 & 684 \\
& Tracktor++V2 \cite{Bergmann2019} & 46.6 & 47.6 & 18.2 & 27.9 & 1,290 \\
& MDP\_SubCNN \cite{Xiang2015} & 47.5 & 55.7 & 30.0 & 18.6 & 628 \\
& CDA\_DDAL \cite{Bae2018} & 51.3 & 54.1 & 36.3 & 22.2 & 544 \\
& MPNTrack \cite{Braso2020} & 51.5 & 58.6 & 31.2 & 25.9 & \textbf{375} \\
& Lif\_T \cite{Hornakova2020} & 52.5 & 60.0 & 33.8 & 25.8 & 1,047 \\
& EAMTT \cite{Matilla2016} & 53.0 & 54.0 & 35.9 & 19.6 & 776 \\
& AP\_HWDPL \cite{aphdwpl} & 53.0 & 52.0 & 29.1 & 20.2 & 708 \\
& NOMTwSDP \cite{Choi2015} & 55.5 & \uul{59.1} & 39.0 & 25.8 & \ul{427} \\
& RAR15 \cite{Fang2018} & \uul{56.5} & \ul{61.3} & \ul{45.1} & \ul{14.6} & \uul{428} \\
& Tube\_TK \cite{pang2020tubetk} & \ul{58.4} & 53.1 & \uul{39.3} & \uul{18.0} & 854 \\
& \textbf{GSDT (Ours)}                   & \bf{60.7}   & \bf{64.6}  & \bf{47.0}     & \bf{10.5} & 480 \\
\midrule

\parbox{2mm}{\multirow{11}{*}{\rotatebox[origin=c]{90}{MOT16}}} 
& MPNTrack \cite{Braso2020} & 58.6 & 61.7 & 27.3 & 34.0 & \bf{354} \\
& Lif\_T \cite{Hornakova2020} & 61.3 & 64.7 & 27.0 & 34.0 & \ul{389} \\
& DeepSORT\_2 \cite{Wojke2017}    & 61.4 & 62.2 & 32.8 & \uul{18.2} & 1,423 \\
& NOMTwSDP16 \cite{Choi2015}  &  62.2 & 62.6 & 32.5 & 31.1 & \uul{406} \\
& VMaxx \cite{8451174}              & 62.6 & 49.2 & 32.7 & 21.1 & 1,389 \\
& RAR16wVGG \cite{Fang2018}            & 63.0 & 63.8 & \ul{39.9} & 22.1 & 482   \\
& TAP \cite{8545450}                  & 64.8 & \bf{73.5} & 38.5 & 21.6 & 571   \\
& CNNMTT \cite{CNNMTT}            & 65.2 & 62.2 & 32.5 & 21.3 & 946   \\    
& POI \cite{yu2016poi}                  & 66.1 & \uul{65.1} & 34.0 & 20.8 & 3,093 \\
& Tube\_TK\_POI \cite{pang2020tubetk} & \uul{66.9} & 62.2 & \uul{39.0} & \bf{16.1} & 1,236 \\
& CTracker\_V1 \cite{peng2020chainedtracker}    & \ul{67.6} & 57.2 & 32.9 & 23.1 & 1,897 \\
& \bf{GSDT (Ours)}                      & \bf{74.5} & \ul{68.1} & \bf{41.2} & \ul{17.3} & 1,229 \\

\midrule
\parbox{2mm}{\multirow{5}{*}{\rotatebox[origin=c]{90}{MOT17}}} 
& MPNTrack \cite{Braso2020} & 58.8 & 61.7 & 28.8 & 33.5 & \textbf{1,185} \\
& Lif\_T \cite{Hornakova2020} & 60.5 & \ul{65.6} & 27.0 & 33.6 & \ul{1,189} \\
& Tube\_TK \cite{pang2020tubetk} & 63.0 & 58.6 & 31.2 & \ul{19.9} & 4,137 \\
& CTrackerV1 \cite{peng2020chainedtracker} & \uul{66.6} & 57.4 & \uul{32.2} & \uul{24.2} & 5,529 \\
& CTTrack17 \cite{zhou2020tracking} & \ul{67.8} & \uul{64.7} & \ul{34.6} & 24.6 & \uul{3,039} \\
& \bf{GSDT (Ours)} & \bf{73.2} & \bf{66.5} & \bf{41.7} & \bf{17.5} & 3,891 \\

\midrule
\parbox{2mm}{\multirow{4}{*}{\rotatebox[origin=c]{90}{\vspace{0.3cm} MOT20}}} & SORT20 \cite{Bewley2016}               & 42.7 & 45.1& 16.7 & \uul{26.2} & 4,334 \\
& Tracktor++V2 \cite{Bergmann2019}                 & \uul{52.6} & \uul{52.7} & \uul{29.4} & 26.7 & \bf{1,648} \\
& MPNTrack \cite{Braso2020}                        & \ul{57.6} & \ul{59.1} & \ul{38.2} & \ul{22.5} & \bf{1,210} \\
& \textbf{GSDT (Ours)}                       & \bf{67.1} & \bf{67.5} & \bf{53.1} & \bf{13.2} & \uul{3,133} \\
\bottomrule
\end{tabular}

\label{tab:mot_test}
\vspace{-0.33cm}
\end{table}
\begin{table}[t]
\centering
\caption{Detection evaluation on MOT17Det/MOT20Det test set.}
\vspace{-0.3cm}
\resizebox{\linewidth}{!}{
\begin{tabular}{@{}clrrrr@{}}
\toprule
& Method & AP(\%)$\uparrow$ & MODA(\%)$\uparrow$ & Recall(\%)$\uparrow$ & Precision(\%)$\uparrow$ \\

\midrule
\parbox{2mm}{\multirow{6}{*}{\rotatebox[origin=c]{90}{MOT17}}} & DPM \cite{Felzenszwalb2009} & 0.61 & 31.2 & 68.1 & 64.8 \\
& FRCNN \cite{Ren2015}        & 0.72 & 68.5 & 77.3 & 89.8 \\
& ZIZOM \cite{Lin2018}        & 0.81 & 72.0 & 83.3 & 88.0 \\
& SDP \cite{Yang2016}         & 0.81 & 76.9 & 83.5 & \bf{92.6} \\
& F\_ViPeD\_B \cite{ciampi2020virtual}         & 0.89 & -14.4 & \bf{93.2} & 46.4 \\
& \textbf{GSDT (Ours)}               & \bf{0.89} & \bf{78.1} & 90.7 & 87.8 \\

\midrule
\parbox{2mm}{\multirow{3}{*}{\rotatebox[origin=c]{90}{MOT20}}} & ViPeD20 \cite{ciampi2020virtual}      & 0.80 & 46.0 & 86.5 & 68.1 \\
\vspace{-0.15cm} \\
& \bf{GSDT (Ours)}         & \bf{0.81} & \bf{79.3} & \bf{88.6} & \bf{90.6} \\
\bottomrule

\end{tabular}
\label{tab:mot17det}
}
\vspace{-0.35cm}
\end{table}
\begin{figure*}
\vspace{1.5mm}
    \centering
    \includegraphics[width=1.0\linewidth]{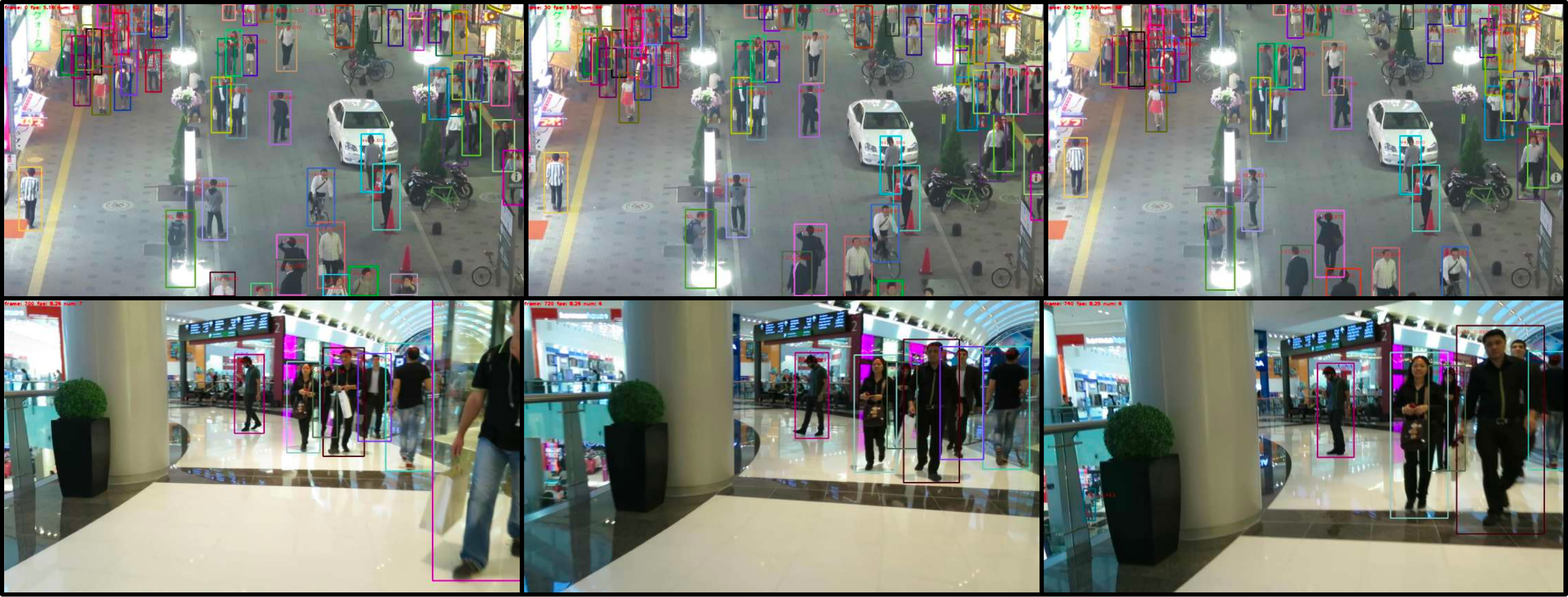}
    \vspace{-0.75cm}
    \caption{Visualization of our detection and tracking results on two sequences of the MOT17 test set.}
    \label{fig:qual_viz}
    \vspace{-0.45cm}
\end{figure*}

\vspace{1.5mm}\noindent\textbf{Evaluating Multi-Object Tracking.}
We summarize results on the MOTChallenges for ours and published methods in Table \ref{tab:mot_test}, marking the \textbf{best}, \ul{second best}, and \uul{third best}. Across most metrics (\emph{e.g.}, primary metrics MOTA and IDF1), we achieve the best performance on all 2DMOT2015/MOT16/MOT17/MOT20 datasets among published methods. As the MOTA metric emphasizes more on detection side, we believe our strong MOTA performance shows that our detector is stronger than prior methods. Also, our strong IDF1 performance suggests that our method is able to produce temporally-consistent tracklets due to the spatial-temporal relation modeling via GNNs. Moreover, when comparing with prior joint MOT methods without modeling object relations such as \cite{B2016, pang2020tubetk}, our method achieves better performance as shown in 2DMOT15/MOT16/MOT17 benchmarks, suggesting that adding object relation modeling is useful in joint MOT. When comparing with prior disjoint MOT methods with relation modeling for data association such as \cite{Braso2020}, our method also performs better across all benchmarks, suggesting that only modeling object relations in data association is not enough, and that it is useful to model object relations in object detection as well. All these results confirm our hypothesis that modeling object relations in the joint MOT framework can lead to better performance. Note that all numbers in Table \ref{tab:mot_test} are updated from the \textcolor{blue}{\href{https://motchallenge.net/results/MOT20/?det=All}{official leaderboard}} by 2021/03/22. To visually verify our results, we also provide qualitative results of our method in Fig. \ref{fig:qual_viz}

\vspace{1mm}\noindent\textbf{Evaluating Object Detection.}
We compare our method with published methods in Table \ref{tab:mot17det}.
We observed that our detector outperforms prior methods in most of the metrics. We highlight that we achieve a higher Average Precision than most of the prior methods, which could be attributed to our use of GNNs that makes it easier to find existing objects and harder to generate false positives by exploiting object relation information.

\section{Ablation Study}
We conduct ablation study on the 2DMOT2015 validation set, removing sequences overlapping with our training set.

\begin{table}[t]
\renewcommand{\arraystretch}{1.3}
\setlength\tabcolsep{1.5pt}
\centering
\caption{Ablation study on effectiveness of GNNs.}
\vspace{-0.3cm}
\resizebox{\linewidth}{!}{
\begin{tabular}{@{}lrrrrrrr@{}}
\toprule
Method & MOTA(\%)$\uparrow$ & IDF1(\%)$\uparrow$ & MT(\%)$\uparrow$ & ML(\%)$\downarrow$ & FP$\downarrow$ & FN$\downarrow$ & IDS$\downarrow$\\
\midrule
No GNNs & 79.5 & 73.8 & \bf{71.8} & \bf{8.8} & 1,948 & \bf{1,487} & 139 \\
1-layer GNNs & 81.6 & \bf{77.4} & 66.3 & 13.5 & 1,394 & 1,733 & 68 \\
2-layer GNNs & 81.9 & 75.0 & 65.7 & 13.5 & \bf{1,234} & 1,866 & \bf{54} \\
3-layer GNNs & \bf{82.1} & 76.4 & 65.4 & 12.6 & 1,281 & 1,793 & 71 \\
\bottomrule
\end{tabular}
\label{tab:ablation_gnn}
}
\vspace{-0.35cm}
\end{table}

\vspace{1mm}\noindent\textbf{Effect of GNNs.}
To answer the question on whether adding a GNN module is useful, we train our model with \{1,2,3\}-layer/no GNNs. The results are shown in Table \ref{tab:ablation_gnn}. When comparing 1-layer/no GNNs model, we see clear improvements in primary metrics MOTA and IDF1, which shows the usefulness of GNNs in the task of MOT. Also, with GNNs, we see a clear decrease in IDS comparing to the no-GNN model, which suggests that GNNs help learn more discriminative features and reduce confusion in data association. Moreover, when comparing \{1,2,3\}-layer GNNs model, we see further improvements in MOTA but lower numbers in IDF1, especially 2-layer model. We hypothesize that adding more layers might have made our model emphasize more on detection but less on temporally-consistent tracking. How to balance the task of detection and data association during the joint MOT learning can be a very interesting research topic in the future. Note that we use AGNNConv \cite{thekumparampil2018attentionbased} in this ablation study and we do not explore more than 3 GNN layers as GPU memory is not enough to fit our model.

\section{Conclusions}
We proposed a new instance of joint MOT approach that simultaneously optimizes object detection and data association modules. Moreover, to model spatial-temporal object relations, we used GNNs to learn more discriminative features that benefit both detection and data association. Through extensive experiments, we showed effectiveness of GNNs under the joint MOT framework and achieved state-of-the-art performance on the MOT challenges. Our code will be released so that future follow-up work can easily build on top of our method to advance the state-of-the-art.


\bibliographystyle{IEEEtran}
\bibliography{IEEEabrv,main}

\end{document}